\documentclass[letterpaper]{article} 
\usepackage{aaai23}  
\usepackage{times}  
\usepackage{helvet}  
\usepackage{courier}  
\usepackage[hyphens]{url}  
\usepackage{graphicx} 
\usepackage{natbib}  
\usepackage{caption} 
\usepackage{algorithm}
\usepackage{algorithmic}

\usepackage{amsfonts}
\usepackage{xcolor}
\usepackage{amssymb}
\usepackage{amsxtra}
\usepackage{multirow}
\usepackage{diagbox}
\usepackage{longtable}
\usepackage{array}
\usepackage{graphicx}
\usepackage{amsmath, bm}
\usepackage{booktabs} 
\usepackage{arydshln}
\usepackage{xspace}
\usepackage{enumitem}
\usepackage{threeparttable}
\usepackage{dsfont}
\usepackage{makecell}
\usepackage{subfigure}
\usepackage{multirow}
\usepackage{tablefootnote}

\newcommand{\model}{RESDSQL}
\newcommand{\smodel}{RESDSQL }
\newcommand{\beq}[1]{\begin{equation}#1\end{equation}}
\newcommand{\beqn}[1]{\begin{eqnarray}#1\end{eqnarray}}

\usepackage{newfloat}
\usepackage{listings}
\DeclareCaptionStyle{ruled}{labelfont=normalfont,labelsep=colon,strut=off} 
\floatstyle{ruled}
\newfloat{listing}{tb}{lst}{}
\floatname{listing}{Listing}
\title{RESDSQL: Decoupling Schema Linking and Skeleton Parsing for Text-to-SQL}
\author {
    Haoyang Li\textsuperscript{\rm 1,\rm 2,\rm 3},
    Jing Zhang\textsuperscript{\rm 1,\rm 2,\rm 3}\thanks{Jing Zhang is the corresponding author.},
    Cuiping Li\textsuperscript{\rm 1,\rm 2,\rm 3},
    Hong Chen\textsuperscript{\rm 1,\rm 2,\rm 3}
}
\affiliations {
    \textsuperscript{\rm 1} Key Laboratory of Data Engineering and Knowledge Engineering of Ministry of Education, Renmin University of China\\
    \textsuperscript{\rm 2}Engineering Research Center of Ministry of Education on Database and BI\\
    \textsuperscript{\rm 3} Information School, Renmin University of China\\
    \{lihaoyang.cs, zhang-jing, licuiping, chong\}@ruc.edu.cn
}

\usepackage{bibentry}

\begin{document}
\maketitle

\begin{abstract}
One of the recent best attempts at Text-to-SQL is the pre-trained language model. Due to the structural property of the SQL queries, the seq2seq model takes the responsibility of parsing both the schema items (\emph{i.e.}, tables and columns) and the skeleton (\emph{i.e.}, SQL keywords). Such coupled targets increase the difficulty of parsing the correct SQL queries especially when they involve many schema items and logic operators. This paper proposes a ranking-enhanced encoding and skeleton-aware decoding framework to decouple the schema linking and the skeleton parsing. Specifically, for a seq2seq encoder-decode model, its encoder is injected by the most relevant schema items instead of the whole unordered ones, which could alleviate the schema linking effort during SQL parsing, and its decoder first generates the skeleton and then the actual SQL query, which could implicitly constrain the SQL parsing. We evaluate our proposed framework on Spider and its three robustness variants: Spider-DK, Spider-Syn, and Spider-Realistic. The experimental results show that our framework delivers promising performance and robustness. Our code is available at https://github.com/RUCKBReasoning/RESDSQL.
\end{abstract}

\section{Introduction}
Relational databases that are used to store heterogeneous data types including text, integer, float, etc., are omnipresent in modern data management systems. 
However, ordinary users usually cannot make the best use of databases because they are not good at translating their requirements to the database language---\emph{i.e.}, the structured query language (SQL). 
To assist these non-professional users in querying the databases, researchers propose the Text-to-SQL task~\citep{tao2018typesql, ruichu2018an}, which aims to automatically translate users' natural language questions into SQL queries. 
At the same time, related benchmarks are becoming increasingly complex, from the single-domain benchmarks such as ATIS~\citep{srinivasan2017learning} and GeoQuery~\citep{john1996learning} to the cross-domain benchmarks such as WikiSQL~\citep{victor2017seq2sql} and Spider~\citep{tao2018spider}. Most of the recent works are done on Spider because it is the most challenging benchmark which involves many complex SQL operators (such as \texttt{GROUP BY}, \texttt{ORDER BY}, and \texttt{HAVING}, etc.) and nested SQL queries.

\begin{figure}[t]
    \centering	\includegraphics[width=0.46\textwidth]{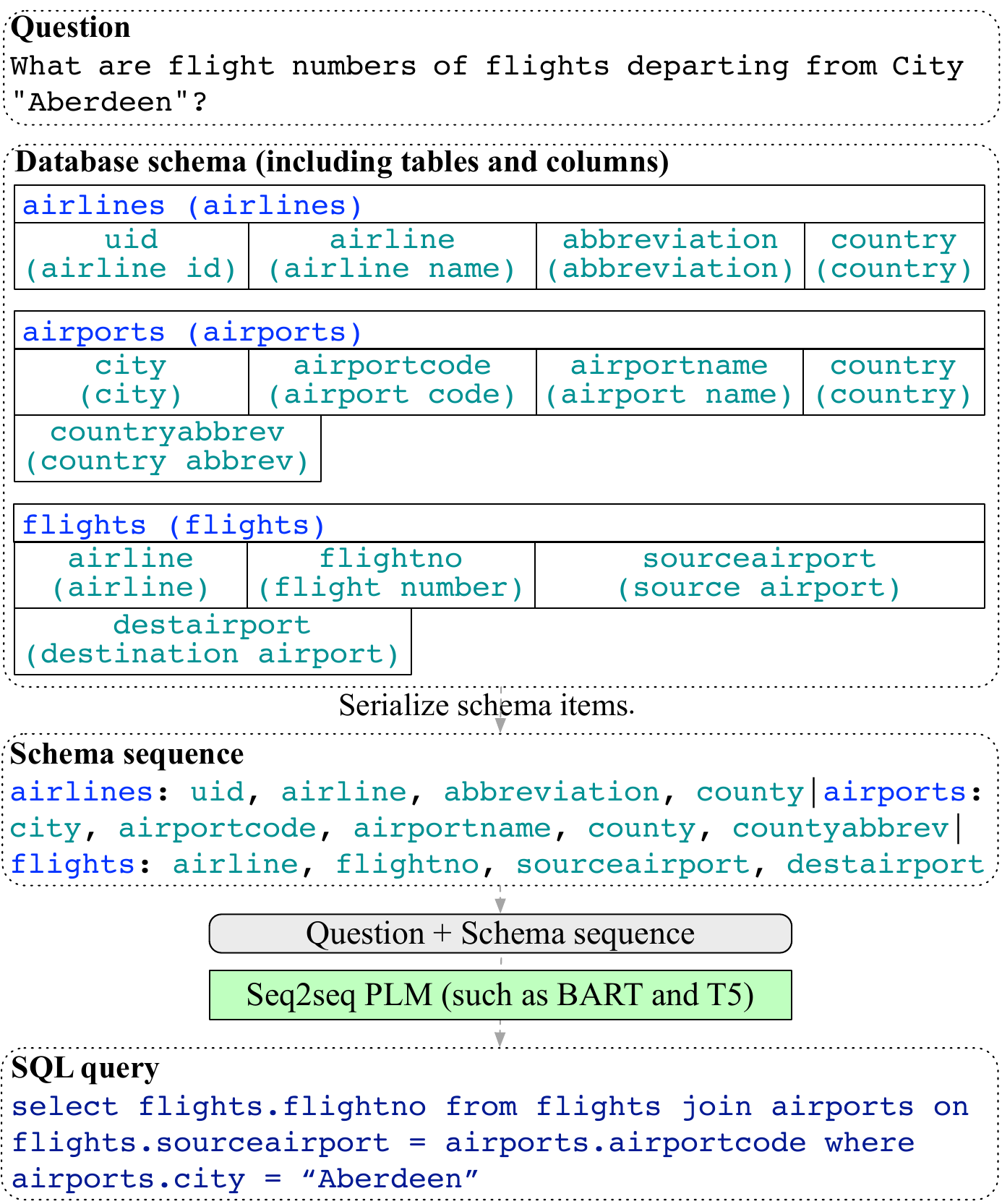}
	\caption{\label{fig:example} Illustration of a Text-to-SQL instance solved by a seq2seq PLM. In the database schema, each schema item is denoted by its ``original name (semantic name)''.}
\end{figure}

With the recent advances in pre-trained language models (PLMs), many existing works formulate the Text-to-SQL task as a semantic parsing problem and use a sequence-to-sequence (seq2seq) model to solve it~\citep{torsten2021picard, peng2021learning, peter2021compositional}. Concretely, as shown in Figure~\ref{fig:example}, given a question and a database schema, the schema items are serialized into a schema sequence where the order of the schema items is either default or random. Then, a seq2seq PLM, such as BART~\citep{mike2020bart} and T5~\citep{colin2020t5}, is leveraged to generate the SQL query based on the concatenation of the question and the schema sequence. We observe that the target SQL query contains not only the skeleton that reveals the logic of the question but also the required schema items. For instance, for a SQL query: ``SELECT petid FROM pets WHERE pet\_age = 1'', its skeleton is ``SELECT \_ FROM \_ WHERE \_'' and its required schema items are ``petid'', ``pets'', and ``pet\_age''. 

Since Text-to-SQL needs to perform not only the schema linking which aligns the mentioned entities in the question to schema items in the database schema, but also the skeleton parsing which parses out the skeleton of the SQL query, the major challenges are caused by a large number of required schema items and the complex composition of operators such as \texttt{GROUP BY}, \texttt{HAVING}, and \texttt{JOIN ON} involved in a SQL query. The intertwining of the schema linking and the skeleton parsing complicates learning even more.

To investigate whether the Text-to-SQL task could become easier if the schema linking and the skeleton parsing are decoupled, we conduct a preliminary experiment on Spider's dev set. Concretely, we fine-tune a T5-Base model to generate the pure skeletons based on the questions (\emph{i.e.}, skeleton parsing task). We observe that the exact match accuracy on such a task achieves about 80\% using the fine-tuned T5-Base. However, even the T5-3B model only achieves about 70\% accuracy~\citep{peter2021compositional, torsten2021picard}. This pre-experiment indicates that decoupling such two objectives could be a potential way of reducing the difficulty of Text-to-SQL.

To realize the above decoupling idea, we propose a \underline{R}anking-enhanced \underline{E}ncoding plus a \underline{S}keleton-aware \underline{D}ecoding framework for Text-to-\underline{SQL} (\model). The former injects a few but most relevant schema items into the seq2seq model's encoder instead of all schema items. In other words, the schema linking is conducted beforehand to filter out most of the irrelevant schema items in the database schema, which can alleviate the difficulty of the schema linking for the seq2seq model. For such purpose, we train an additional cross-encoder to classify the tables and columns simultaneously based on the input question, and then rank and filter them according to the classification probabilities to form a ranked schema sequence. The latter does not add any new modules but simply allows the seq2seq model's decoder to first generate the SQL skeleton, and then the actual SQL query. Since skeleton parsing is much easier than SQL parsing, the first generated skeleton could implicitly guide the subsequent SQL parsing via the masked self-attention mechanism in the decoder.

\paragraph{Contributions} 
(1) We investigate a potential way of decoupling the schema linking and the skeleton parsing to reduce the difficulty of Text-to-SQL. Specifically, we propose a ranking-enhanced encoder to alleviate the effort of the schema linking and a skeleton-aware decoder to implicitly guide the SQL parsing by the skeleton. (2) We conduct extensive evaluation and analysis and show that our framework not only achieves the new state-of-the-art (SOTA) performance on Spider but also exhibits strong robustness. 

\section{Problem Definition}
\paragraph{Database Schema}
A relational database is denoted as $\mathcal{D}$.
The database schema $\mathcal{S}$ of $\mathcal{D}$ includes (1) a set of $N$ tables $\mathcal{T}=\{t_1,t_2,\cdots,t_{N}\}$,  
(2) a set of columns $\mathcal{C}=\{c_1^{1}, \cdots,c_{n_1}^{1},c_1^{2}, \cdots, c_{n_2}^{2},\cdots,c_1^{N},\cdots,c_{n_N}^{N}\}$ associated with the tables, where $n_i$ is the number of columns in the $i$-th table,
(3) and a set of foreign key relations $\mathcal{R} = \{(c^i_k, c^j_h) | c^i_k, c^j_h \in \mathcal{C}\}$, where each $(c^i_k, c^j_h)$ denotes a foreign key relation between column $c^i_k$ and column $c^j_h$. We use $M=\sum_{i=1}^{N}n_i$ to denote the total number of columns in $\mathcal{D}$. 

\paragraph{Original Name and Semantic Name}
We use ``schema items'' to uniformly refer to tables and columns in the database. Each schema item can be represented by an original name and a semantic name. The semantic name can indicate the semantics of the schema item more precisely. As shown in Figure~\ref{fig:example}, it is obvious that the semantic names ``airline id'' and ``destination airport'' are more clear than their original names ``uid'' and ``destairport''. Sometimes the semantic name is the same as the original name.

\paragraph{Text-to-SQL Task}
Formally, given a question $q$ in natural language and a database $\mathcal{D}$ with its schema $\mathcal{S}$, the Text-to-SQL task aims to translate $q$ into a SQL query $l$ that can be executed on $\mathcal{D}$ to answer the question $q$.

\section{Methodology}
In this section, we first give an overview of the proposed framework and then delve into its design details.

\begin{figure*}[t]
    \centering
	\includegraphics[width=0.85\textwidth]{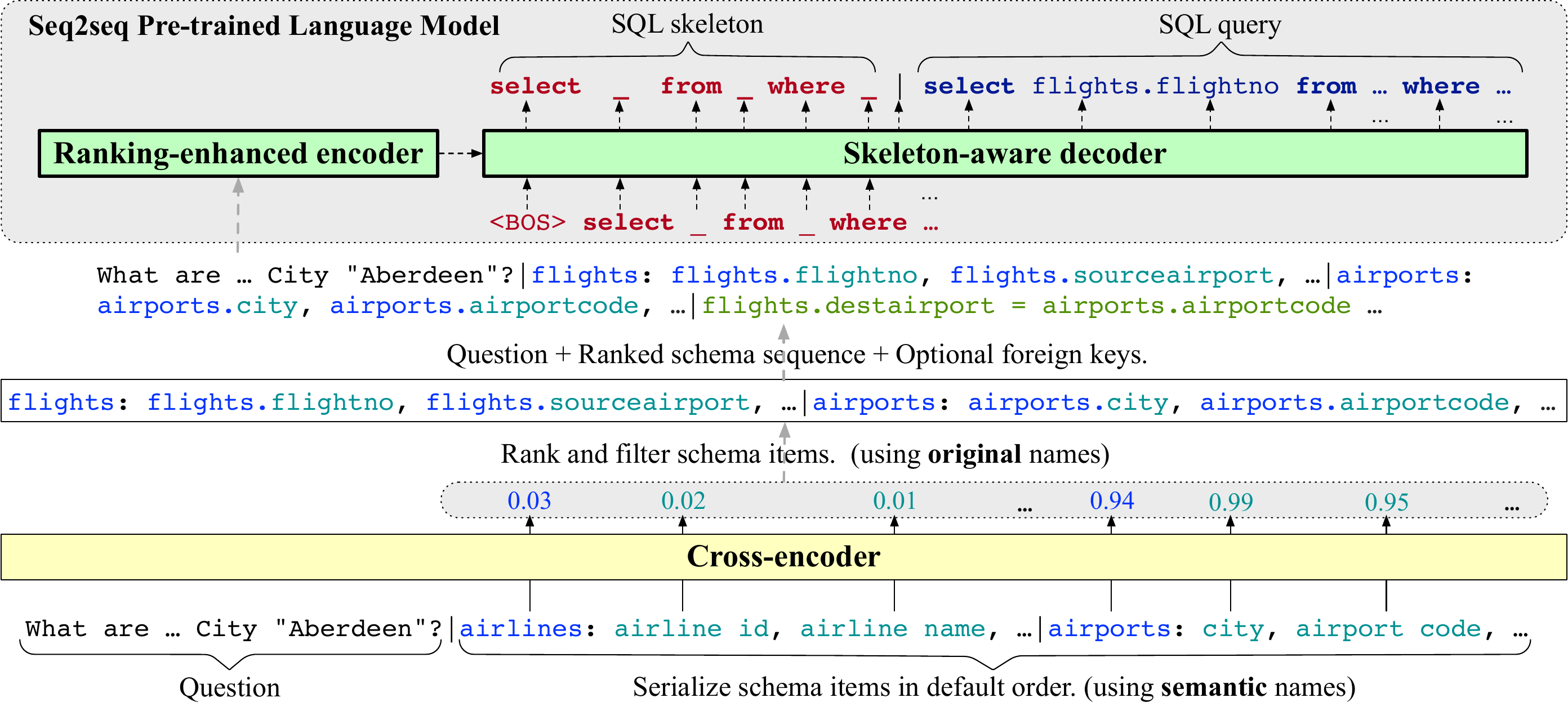}
	\caption{\label{fig:overview} An overview of the ranking-enhanced encoding and skeleton-aware decoding framework. We train a cross-encoder for classifying the schema items. Then we take the question, the ranked schema sequence, and optional foreign keys as the input of the ranking-enhanced encoder. The skeleton-aware decoder first decodes the SQL skeleton and then the actual SQL query.}
\end{figure*}

\subsection{Model Overview}
Following~\citet{peter2021compositional, torsten2021picard}, we treat Text-to-SQL as a translation task, which can be solved by an encoder-decoder transformer model. 
Facing the above problems, we extend the existing seq2seq Text-to-SQL methods by injecting the most relevant schema items in the input sequence and the SQL skeleton in the output sequence, which results in a ranking-enhanced encoder and a skeleton-aware decoder.
We provide the high-level overview of the proposed \smodel framework in Figure~\ref{fig:overview}. The encoder of the seq2seq model receives the ranked schema sequence, such that the schema linking effort could be alleviated during SQL parsing. To obtain such a ranked schema sequence, an additional cross-encoder is proposed to classify the schema items according to the given question, and then we rank and filter them based on the classification probabilities. The decoder of the seq2seq model first parses out the SQL skeleton and then the actual SQL query, such that the SQL generation can be implicitly constrained by the previously parsed skeleton. By doing this, to a certain extent, the schema linking and the skeleton parsing are not intertwined but decoupled.

\subsection{Ranking-Enhanced Encoder}
Instead of injecting all schema items, we only consider the most relevant schema items in the input of the encoder. 
For this purpose, we devise a cross-encoder to classify the tables and columns simultaneously and then rank them based on their probabilities. Based on the ranking order, on one hand, we filter out the irrelevant schema items. On the other hand, we use the ranked schema sequence instead of the unordered schema sequence, so that the seq2seq model could capture potential position information for schema linking. 

As for the input of the cross-encoder, we flatten the schema items into a schema sequence in their \textit{default} order and concatenate it with the question to form an input sequence: 
$X = q\ | \  t_1:c_1^1,\cdots,c^1_{n_1}\ | \cdots |\ t_{N}:c_{1}^{N},\cdots,c_{n_N}^{N}$, where $|$ is the delimiter. To better represent the semantics of schema items, instead of their original names, we use their semantic names which are closer to the natural expression.

\paragraph{Encoding Module}
We feed $X$ into RoBERTa~\citep{yinhan2019roberta}, an improved version of BERT~\cite{jacob2019bert}. Since each schema item will be tokenized into one or more tokens by PLM's tokenizer (\emph{e.g.}, the column ``airline id'' will be split into two tokens: ``airline'' and ``id''), and our target is to represent each schema item as a whole for classification, we need to pool the output embeddings belonging to each schema item. 
To achieve this goal, we use a pooling module that consists of a two-layer BiLSTM~\citep{sepp1997long} and a non-linear fully-connected layer. After pooling, each table embedding can be denoted by $\boldsymbol{T}_{i} \in \mathbb{R}^{1 \times d}\ (i \in \{1, ..., N\})$ and each column embedding can be denoted by $\boldsymbol{C}_{k}^{i} \in \mathbb{R}^{1 \times d}\ (i \in \{1, ..., N\}, k \in \{1, ..., n_{i}\})$, where $d$ denotes the hidden size.

\paragraph{Column-Enhanced Layer} 
We observe that some questions only mention the column name rather than the table name. 
For example in Figure~\ref{fig:example}, the question mentions a column name ``city'', but its corresponding table name ``airports'' is ignored. This table name missing issue may compromise the table classification performance. Therefore, we propose a column-enhanced layer to inject the column information into the corresponding table embedding. In this way, a table could be identified even if the question only mentions its columns. Concretely, for the $i$-th table, we inject the column information $\boldsymbol{C}_{:}^{i} \in \mathbb{R}^{n_i\times d}$ into the table embedding $\boldsymbol{T}_i$ by stacking a multi-head scaled dot-product attention layer~\citep{ashish2017attention} and a feature fusion layer on the top of the encoding module:
\begin{align}
\label{eq:attn_c} &\boldsymbol{T}_{i}^{C} = MultiHeadAttn(\boldsymbol{T}_{i}, \boldsymbol{C}_{:}^{i}, \boldsymbol{C}_{:}^{i}, h), \\ \nonumber
&\hat{\boldsymbol{T}}_i = Norm(\boldsymbol{T}_i+\boldsymbol{T}_{i}^{C}).
\end{align}
\noindent Here, $\boldsymbol{T}_{i}$ acts as the query and $\boldsymbol{C}_{:}^{i}$ acts as both the key and the value, $h$ is the number of heads, and $Norm(\cdot)$ is a row-wise $L_{2}$ normalization function. $\boldsymbol{T}_{i}^{C}$ represents the column-attentive table embedding. We fuse the original table embedding $\boldsymbol{T}_{i}$ and the column-attentive table embedding $\boldsymbol{T}_{i}^C$ to obtain the column-enhanced table embedding $\hat{\boldsymbol{T}}_i \in \mathbb{R}^{1\times d}$.

\paragraph{Loss Function of Cross-Encoder}
Cross-entropy loss is a well-adopted loss function in classification tasks. However, since a SQL query usually involves only a few tables and columns in the database, the label distribution of the training set is highly imbalanced. As a result, the number of negative examples is many times that of positive examples, which will induce serious training bias. To alleviate this issue, we employ the focal loss~\citep{tsung2017focal} as our classification loss. Then, we form the loss function of the cross-encoder in a multi-task learning way, which consists of both the table classification loss and the column classification loss, \emph{i.e.}, 
\begin{equation}
\begin{aligned}
    &\mathcal{L}_1 = \frac{1}{N}\sum_{i=1}^{N} FL(y_{i}, \hat{y}_i) +  \frac{1}{M}\sum_{i=1}^{N} \sum_{k=1}^{n_i} FL( y^{i}_{k}, \hat{y}^i_{k}),\\
\end{aligned}
\end{equation}

\noindent where $FL$ denotes the focal loss function and $y_{i}$ is the ground truth label of the $i$-th table. $y_{i} = 1$ indicates the table is referenced by the SQL query and 0 otherwise. $y_{k}^i$ is the ground truth label of the $k$-th column in the $i$-th table. Similarly, $y_{k}^i = 1$ indicates the column is referenced by the SQL query and 0 otherwise. $\hat{y}_i$ and $\hat{y}_k^i$ are predicted probabilities, which are estimated by two different MLP modules based on the table and column embeddings $\hat{\boldsymbol{T}}_{i}$ and $\boldsymbol{C}_{k}^{i}$:
\beqn{
    \hat{y}_i &=& \sigma((\hat{\boldsymbol{T}}_{i}\boldsymbol{U}^t_{1}+\boldsymbol{b}^t_{1})\boldsymbol{U}^t_{2}+\boldsymbol{b}^t_{2}), \\ \nonumber
    \hat{y}_k^i &=& \sigma((\boldsymbol{C}_{k}^{i}\boldsymbol{U}^c_{1}+\boldsymbol{b}^c_{1})\boldsymbol{U}^c_{2}+\boldsymbol{b}^c_{2}),
}

\noindent where $\boldsymbol{U}^t_{1}$, $\boldsymbol{U}^c_{1} \in \mathbb{R}^{d\times w}$, $\boldsymbol{b}^t_{1}$, $\boldsymbol{b}^c_{1} \in \mathbb{R}^{w}$, $\boldsymbol{U}^t_{2}$, $\boldsymbol{U}^c_{2} \in \mathbb{R}^{w\times 2}$, $\boldsymbol{b}^t_{2}$, $\boldsymbol{b}^c_{2} \in \mathbb{R}^{2}$ are trainable parameters, and $\sigma(\cdot)$ denotes Softmax.

\paragraph{Prepare Input for Ranking-Enhanced Encoder}
During inference, for each Text-to-SQL instance, we leverage the above-trained cross-encoder to compute a probability for each schema item. Then, we only keep top-$k_{1}$ tables in the database and top-$k_{2}$ columns for each remained table to form a ranked schema sequence. $k_1$ and $k_2$ are two important hyper-parameters. When $k_1$ or $k_2$ is too small, a portion of the required tables or columns may be excluded, which is fatal for the subsequent seq2seq model. As $k_1$ or $k_2$ becomes larger, more and more irrelevant tables or columns may be introduced as noise. Therefore, we need to choose appropriate values for $k_1$ and $k_2$ to ensure a high recall while preventing the introduction of too much noise. The input sequence for the ranking-enhanced encoder (\emph{i.e.}, seq2seq model's encoder) is formed as the concatenation of the question, the ranked schema sequence, and optional foreign key relations (see Figure~\ref{fig:overview}). Foreign key relations contain rich information about the structure of the database, which could promote the generation of the \texttt{JOIN ON} clauses. In the ranked schema sequence, we use the original names instead of the semantic names because the schema items in the SQL queries are represented by their original names, and using the former will facilitate the decoder to \textit{directly copy required schema items} from the input sequence.

\subsection{Skeleton-Aware Decoder}
Most seq2seq Text-to-SQL methods tell the decoder to generate the target SQL query directly. However, the apparent gap between the natural language and the SQL query makes it difficult to perform the correct generation. To alleviate this problem, we would like to decompose the SQL generation into two steps: (1) generate the SQL skeleton based on the semantics of the question, and then (2) select the required ``data'' (\emph{i.e.}, tables, columns, and values) from the input sequence to fill the slots in the skeleton. 

To realize the above decomposition idea without adding additional modules, we propose a new generation objective based on the intrinsic characteristic of the transformer decoder, which generates the $t$-th token depending on not only the output of the encoder but also the output of the decoder before the $t$-th time step~\cite{ashish2017attention}. Concretely, instead of decoding the target SQL directly, we encourage the decoder to first decode the skeleton of the SQL query, and based on this, we continue to decode the SQL query.

By parsing the skeleton first and then parsing the SQL query, at each decoding step, SQL generation will be easier because the decoder could either copy a ``data'' from the input sequence or a SQL keyword from the previously parsed skeleton. Now, the objective of the seq2seq model is:
\beq{
\mathcal{L}_2 =  \frac{1}{G}\sum_{i=1}^G p(l^{s}_i,l_i|S_i),
}

\noindent where $G$ is the number of Text-to-SQL instances, $S_i$ is the input sequence of the $i$-th instance which consists of the question, the ranked schema sequence, and optional foreign key relations. $l_i$ denotes the $i$-th target SQL query and $l^{s}_i$ is the skeleton extracted from $l_i$. We will present some necessary details on how to normalize SQL queries and how to extract their skeletons.

\paragraph{SQL Normalization}
The Spider dataset is manually created by 11 annotators with different annotation habits, which results in slightly different styles among the final annotated SQL queries, such as uppercase versus lowercase keywords. Although different styles have no impact on the execution results, the model requires some extra effort to learn and adapt to them. To reduce the learning difficulty, we normalize the original SQL queries before training by (1) unifying the keywords and schema items into lowercase, (2) adding spaces around parentheses and replacing double quotes with single quotes, (3) adding an \texttt{ASC} keyword after the \texttt{ORDER BY} clause if it does not specify the order, and (4) removing the \texttt{AS} clause and replacing all table aliases with their original names. We present an example in Table~\ref{tab:sqls}.

\paragraph{SQL Skeleton Extraction}
Based on the normalized SQL queries, we can extract their skeletons which only contain SQL keywords and slots. Specifically, given a normalized SQL query, we keep its keywords and replace the rest parts with slots. Note that we do not keep the \texttt{JOIN ON} keyword because it is difficult to find a counterpart from the question~\cite{yujian2021naturalsql}. As shown in Table~\ref{tab:sqls}, although the original SQL query looks complex, its skeleton is simple and each keyword can find a counterpart from the question. For example, ``order by \_ asc'' in the skeleton can be inferred from ``ordered by title?'' in the question.

\begin{table}[]
    \newcolumntype{?}{!{\vrule width 1pt}}
    \newcolumntype{C}{>{\arraybackslash}p{6.5cm}}
    \centering
    \small
    \begin{tabular}{l?C}
    \toprule
        \multirow{2}{*}{Q} & List the duration, file size and format of songs whose genre is pop, ordered by title? \\ \hline
        \multirow{4}{*}{SQL$_{o}$} & SELECT T1.duration,  T1.file\_size, T1.formats FROM files AS T1 JOIN song AS T2 ON T1.f\_id  =  T2.f\_id WHERE T2.genre\_is = ``pop'' ORDER BY T2.song\_name\\\hline
        \multirow{3}{*}{SQL$_{n}$} & select files.duration, files.file\_size, files.formats from files join song on files.f\_id = song.f\_id where song.genre\_is = `pop' order by song.song\_name asc\\\hline
        SQL$_{s}$ & select \_ from \_ where \_ order by \_ asc\\    
    \bottomrule
    \end{tabular}
    \caption{An example from Spider. Here, Q, SQL$_{o}$, SQL$_{n}$, and SQL$_{s}$ denote the question, the original SQL query, the normalized SQL query, and the SQL skeleton, respectively.}
    \label{tab:sqls}
\end{table}

\paragraph{Execution-Guided SQL Selector}
Since we do not constrain the decoder with SQL grammar, the model may generate some illegal SQL queries. To alleviate this problem, we follow \citet{alane2020exploring} to use an execution-guided SQL selector which performs the beam search during the decoding procedure and then selects the first executable SQL query in the beam as the final result.

\section{Experiments}
\begin{table*}[]
    \newcolumntype{?}{!{\vrule width 1pt}}
    \centering
    \small
    \begin{tabular}{l?cc?cc}
    \toprule
        \multirow{2}*{Approach} &  \multicolumn{2}{c?}{Dev Set} & \multicolumn{2}{c}{Test Set} \\
          & \textbf{EM} & \textbf{EX} & \textbf{EM} & \textbf{EX} \\

        \midrule
        \multicolumn{5}{c}{\textbf{Non-seq2seq methods}} \\
        \midrule

        RAT-SQL + \textsc{Grappa} \citep{tao2021grappa} & 73.4 & - & 69.6 & - \\
        RAT-SQL + GAP + NatSQL \citep{yujian2021naturalsql} & 73.7 & 75.0 & 68.7 & 73.3 \\
        \textsc{SmBoP} + \textsc{Grappa} \citep{ohad2021smbop}& 74.7 & 75.0 & 69.5 & 71.1 \\
        DT-Fixup SQL-SP + RoBERTa \citep{peng2021optimizing} & 75.0 & - & 70.9 & - \\
        LGESQL + ELECTRA \citep{ruisheng2021lgesql}& 75.1 & - & \textbf{72.0} & - \\
        S$^{2}$SQL + ELECTRA \citep{binyuan2022s2sql} & 76.4 & - & \textbf{72.1} & - \\

        \midrule
        \multicolumn{5}{c}{\textbf{Seq2seq methods}} \\
        \midrule

        T5-3B \citep{torsten2021picard} & 71.5 & 74.4 & 68.0 & 70.1 \\
        T5-3B + PICARD \citep{torsten2021picard} & 75.5 & 79.3 & \textbf{71.9} & 75.1 \\
        RASAT + PICARD \citep{jiexing2022rasat} & 75.3 & 80.5 & 70.9 & 75.5 \\

        \midrule
        \multicolumn{5}{c}{\textbf{Our proposed method}} \\
        \midrule

        \model-Base & 71.7 & 77.9 & - & - \\
        \model-Base + NatSQL & 74.1 & 80.2 & - & - \\
        \model-Large & 75.8 & 80.1 & - & - \\
        \model-Large + NatSQL & 76.7 & 81.9 & - & - \\
        \model-3B & 78.0 & 81.8 & - & - \\
        \model-3B + NatSQL & \textbf{80.5} & \textbf{84.1} & \textbf{72.0} & \textbf{79.9} \\
    \bottomrule
    \end{tabular}
    \caption{EM and EX results on Spider's development set and hidden test set (\%). We compare our approach with some powerful baseline methods from the top of the official leaderboard of Spider.}
    \label{tab:main_results}
\end{table*}

\subsection{Experimental Setup}
\paragraph{Datasets}
We conduct extensive experiments on Spider and its three variants which are proposed to evaluate the robustness of the Text-to-SQL parser. Spider~\citep{tao2018spider} is the most challenging benchmark for the cross-domain and multi-table Text-to-SQL task. 
Spider contains a training set with 7,000 samples\footnote{Spider also provides additional 1,659 training samples, which are collected from some single-domain datasets, such as GeoQuery~\citep{john1996learning} and Restaurants~\citep{alessandra2012automatic}. But following \citep{torsten2021picard}, we ignore this part in our training set.}, a dev set with 1,034 samples, and a hidden test set with 2,147 samples. There is no overlap between the databases in different splits. For robustness, we train the model on Spider's training set but evaluate it on Spider-DK~\citep{yujian2021exploring} with 535 samples, Spider-Syn~\citep{yujian2021towards} with 1034 samples, and Spider-Realistic~\citep{xiang2021structure} with 508 samples. These evaluation sets are derived from Spider by modifying questions to simulate real-world application scenarios. Concretely, Spider-DK incorporates some domain knowledge to paraphrase questions. Spider-Syn replaces schema-related words with synonyms in questions. Spider-Realistic removes explicitly mentioned column names in questions.
 
\paragraph{Evaluation Metrics} 
To evaluate the performance of the Text-to-SQL parser, following \citealp{tao2018spider, ruiqi2020semantic}, we adopt two metrics: Exact-set-Match accuracy (EM) and EXecution accuracy (EX). The former measures whether the predicted SQL query can be exactly matched with the gold SQL query by converting them into a special data structure~\citep{tao2018spider}. The latter compares the execution results of the predicted SQL query and the gold SQL query. The EX metric is sensitive to the generated values, but the EM metric is not. In practice, we use the sum of EM and EX to select the best checkpoint of the seq2seq model. For the cross-encoder, we use Area Under ROC Curve (AUC) to evaluate its performance. Since the cross-encoder classifies tables and columns simultaneously, we adopt the sum of table AUC and column AUC to select the best checkpoint of the cross-encoder.

\paragraph{Implementation Details} 
We train \smodel in two stages. In the first stage, we train the cross-encoder for ranking schema items. The number of heads $h$ in the column-enhanced layer is 8. 
We use AdamW~\citep{ilya2019decoupled} with batch size 32 and learning rate 1e-5 for optimization. 
In the focal loss, the focusing parameter $\gamma$ and the weighted factor $\alpha$ are set to 2 and 0.75 respectively. Then, $k_1$ and $k_2$ are set to 4 and 5 according to the statistics of the datasets.  
For training the seq2seq model in the second stage, we consider three scales of T5: Base, Large, and 3B. We fine-tune them with Adafactor~\citep{noam2018adafactor} using different batch size (bs) and learning rate (lr), resulting in \model-Base (bs = 32, lr = 1e-4), \model-Large (bs = 32, lr = 5e-5), and \model-3B (bs = 96, lr = 5e-5). For both stages of training, we adopt linear warm-up (the first 10\% training steps) and cosine decay to adjust the learning rate. We set the beam size to 8 during decoding. Moreover, following~\citet{xi2020bridging}, we extract potentially useful contents from the database to enrich the column information.

\paragraph{Environments} We conduct all experiments on a server with one NVIDIA A100 (80G) GPU, one Intel(R) Xeon(R) Silver 4316 CPU, 256 GB memory and Ubuntu 20.04.2 LTS operating system.

\subsection{Results on Spider}
Table~\ref{tab:main_results} reports EM and EX results on Spider. Noticeably, we observe that \model-Base achieves better performance than the bare T5-3B, which indicates that our decoupling idea can substantially reduce the learning difficulty of Text-to-SQL. Then, \model-3B outperforms the best baseline by 1.6\% EM and 1.3\% EX on the dev set. Furthermore, when combined with NatSQL~\citep{yujian2021naturalsql}, an intermediate representation of SQL, \model-Large achieves competitive results compared to powerful baselines on the dev set, and \model-3B achieves new SOTA performance on both the dev set and the test set. Specifically, on the dev set, \model-3B + NatSQL brings 4.2\% EM and 3.6\% EX absolute improvements. On the hidden test set, \model-3B + NatSQL achieves competitive performance on EM and dramatically increases EX from 75.5\% to 79.9\% (+4.4\%), showing the effectiveness of our approach. The reason for the large gap between EM (72.0\%) and EX (79.9\%) is that EM is overly strict~\citep{ruiqi2020semantic}. For example in Spider, given a question ``Find id of the candidate who most recently accessed the course?'', its gold SQL query is ``select candidate\_id from candidate\_assessments order by assessment\_date desc limit 1''. In fact, there is another SQL query ``select candidate\_id from candidate\_assessments where assessment\_date = (select max(assessment\_date) from candidate\_assessments)'' which can also be executed to answer the question (\emph{i.e.}, EX is positive). However, EM will judge the latter to be wrong, which leads to false negatives.

\subsection{Results on Robustness Settings}
Recent studies~\citep{yujian2021towards, xiang2021structure} show that neural Text-to-SQL parsers are fragile to question perturbations because explicitly mentioned schema items are removed or replaced with semantically consistent words (\emph{e.g.}, synonyms), which increases the difficulty of schema linking. Therefore, more and more efforts have been recently devoted to improving the robustness of neural Text-to-SQL parsers, such as TKK~\citep{bowen2022sun} and S\textsc{un}~\citep{chang2022towards}. To validate the robustness of \model, we train our model on Spider's training set and evaluate it on three challenging Spider variants: Spider-DK, Spider-Syn, and Spider-Realistic. Results are reported in Table~\ref{tab:robustness}. We can observe that in all three datasets, \model-3B + NatSQL surprisingly outperforms all strong competitors by a large margin, which suggests that our decoupling idea can also improve the robustness of seq2seq Text-to-SQL parsers. We attribute this to the fact that our proposed cross-encoder can alleviate the difficulty of schema linking and thus exhibits robustness in terms of question perturbations.

\subsection{Ablation Studies}
We take a thorough ablation study on Spider's dev set to analyze the effect of each design.

\paragraph{Effect of Column-Enhanced Layer} We investigate the effectiveness of the column-enhanced layer, which is designed to alleviate the table missing problem. Table~\ref{tab:ablation_1st_stage} shows that removing such a layer will lead to a decrease in the total AUC, as it can inject the human prior into the cross-encoder.

\paragraph{Effect of Focal Loss} We also study the effect of focal loss by replacing it with the cross-entropy loss for schema item classification. Table~\ref{tab:ablation_1st_stage} shows that cross-entropy leads to a performance drop because it cannot alleviate the label-imbalance problem in the training data. 

\paragraph{Effect of Ranking Schema Items} As shown in Table~\ref{tab:ablation_2nd_stage}, when we replace the ranked schema sequence with the original unordered schema sequence, EM and EX significantly decrease by 4.5\% and 7.8\% respectively. This result proves that the ranking-enhanced encoder takes a crucial role.

\paragraph{Effect of Skeleton Parsing} Meanwhile, from Table~\ref{tab:ablation_2nd_stage}, we can observe that EM and EX drop 0.7\% and 0.8\% respectively when removing the SQL skeleton from the decoder's output (\emph{i.e.}, without skeleton parsing). This is because the seq2seq model needs to make extra efforts to bridge the gap between natural language questions and SQL queries when parsing SQL queries directly.

\begin{table*}[htbp]
    \centering
    \small
    \begin{tabular}{lcccccc}
    \toprule
        \multirow{2}*{Approach} & \multicolumn{2}{c}{Spider-DK} & \multicolumn{2}{c}{Spider-Syn} & \multicolumn{2}{c}{Spider-Realistic} \\
         & \textbf{EM} & \textbf{EX} & \textbf{EM} & \textbf{EX} & \textbf{EM} & \textbf{EX} \\
        \midrule
        RAT-SQL + BERT~\citep{bailin2020ratsql} & 40.9 & - & 48.2 & - & 58.1 & 62.1 \\
        RAT-SQL + \textsc{Grappa}~\citep{tao2021grappa} & 38.5 & - & 49.1 & - & 59.3 & - \\
        T5-3B~\citep{chang2022towards} & - & - & 59.4 & 65.3 & 63.2 & 65.0 \\
        LGESQL + ELECTRA~\cite{ruisheng2021lgesql} & 48.4 & - & 64.6 & - & 69.2 & - \\
        TKK-3B~\citep{chang2022towards} & - & - & 63.0 & 68.2 & 68.5 & 71.1 \\
        T5-3B + PICARD~\citep{jiexing2022rasat} & - & - & - & - & 68.7 & 71.4 \\
        RASAT + PICARD~\citep{jiexing2022rasat} & - & - & - & - & 69.7 & 71.9 \\
        LGESQL + ELECTRA + S\textsc{un}~\cite{bowen2022sun} & 52.7 & - & 66.9 & - & 70.9 & - \\
        \midrule
        \model-3B + NatSQL & \textbf{53.3} & \textbf{66.0} & \textbf{69.1} & \textbf{76.9} & \textbf{77.4} & \textbf{81.9} \\
    \bottomrule
    \end{tabular}
    \caption{Evaluation results on Spider-DK, Spider-Syn, and Spider-Realistic (\%).}
    \label{tab:robustness}
\end{table*}

\begin{table}[]
\newcolumntype{?}{!{\vrule width 1pt}}
    \centering
    \small
    \begin{tabular}{@{}l@{ }?@{ }ccc@{}}
    \toprule
        Model variant & \textbf{Table AUC} & \textbf{Column AUC} & \textbf{Total}\\
        \midrule
        Cross-encoder & \textbf{0.9973} & \textbf{0.9957} & \textbf{1.9930} \\
        \quad - w/o enh. layer& 0.9965 & 0.9939 & 1.9904\\
        \quad - w/o focal loss & 0.9958 & 0.9943 & 1.9901\\
    \bottomrule
    \end{tabular}
    \caption{Ablation studies of the cross-encoder.}
    \label{tab:ablation_1st_stage}
\end{table}

\begin{table}[]
\newcolumntype{?}{!{\vrule width 1pt}}
    \centering
    \small
    \begin{tabular}{@{}l@{ }?@{ }cc@{}}
    \toprule
        Model variant & \textbf{EM} (\%) & \textbf{EX} (\%)\\
        \midrule
        \model-Base & \textbf{71.7} & \textbf{77.9} \\
        \quad - w/o ranking schema items & 67.2 & 70.1 \\
        \quad - w/o skeleton parsing & 71.0 & 77.1 \\
    \bottomrule
    \end{tabular}
    \caption{The effect of key designs.}
    \label{tab:ablation_2nd_stage}
\end{table}

\section{Related Work}
Our method is related to the encoder-decoder architecture designed for Text-to-SQL, the schema item classification task, and the intermediate representation.

\subsection{Encoder-Decoder Architecture}
The encoder aims to jointly encode the question and database schema, which is generally divided into sequence encoder and graph encoder. The decoder aims to generate the SQL queries based on the output of the encoder. Due to the special format of SQL, grammar- and execution-guided decoders are studied to constrain the decoding results.

\paragraph{Sequence Encoder} The input is a sequence that concatenates the question with serialized database schema~\citep{tao2021grappa, xi2020bridging}. Then, each token in the sequence is encoded by a PLM encoder, such as BERT~\citep{jacob2019bert} and encoder part of T5~\citep{colin2020t5}.

\paragraph{Graph Encoder} The input is one or more heterogeneous graphs~\citep{bailin2020ratsql, binyuan2022s2sql, ruisheng2021lgesql, ruichu2021sadga}, where a node represents a question token, a table or a column, and an edge represents the relation between two nodes. Then, relation-aware transformer networks~\citep{peter2018self-attention} or relational graph neural networks, such as RGCN~\citep{michael2018modeling} and RGAT~\citep{kai2020relational}, are applied to encode each node. Some works also employ PLM encoders to initialize the representation of nodes on the graph~\citep{ruisheng2021lgesql, bailin2020ratsql, ohad2021smbop}. It is undeniable that the graph encoder can flexibly and explicitly represent the relations between any two nodes via edges (\emph{e.g.}, foreign key relations). However, compared to PLMs, graph neural networks (GNNs) usually cannot be designed too deep due to the limitation of the over-smoothing issue~\citep{deli2020measuring}, which restricts the representation ability of GNNs. Then, PLMs have already encoded language patterns in their parameters after pre-training~\citep{yian2021when}, however, the parameters of GNNs are usually randomized. Moreover, the graph encoder relies heavily on the design of relations, which may limit its robustness and generality on other datasets~\cite{chang2022towards}.

\paragraph{Grammar-Based Decoder} To inject the SQL grammar into the decoder, \citet{pengcheng2017a, jayant2017neural} propose a top-down decoder to generate a sequence of pre-defined actions that can describe the grammar tree of the SQL query. \citet{ohad2021smbop} devise a bottom-up decoder instead of the top-down paradigm. PICARD~\citep{torsten2021picard} incorporates an incremental parser into the auto-regressive decoder of PLMs to prune the invalid partially generated SQL queries during beam search.

\paragraph{Execution-Guided Decoder} Some works use an off-the-shelf SQL executor such as SQLite to ensure grammatical correctness. \citet{chenglong2018robust} leverage a SQL executor to check and discard the partially generated SQL queries which raise errors during decoding. To avoid modifying the decoder, \citet{alane2020exploring} check the executability of each candidate SQL query, which is also adopted by our method.

\subsection{Schema Item Classification}
Schema item classification is often introduced as an auxiliary task to improve the schema linking performance for Text-to-SQL. For example, \textsc{Grappa}~\citep{tao2021grappa} and GAP~\citep{peng2021learning} further pre-train the PLMs by using the schema item classification task as one of the pre-training objectives. Then, Text-to-SQL can be viewed as a downstream task to be fine-tuned. \citet{ruisheng2021lgesql} combine the schema item classification task with the Text-to-SQL task in a multi-task learning way.
The above-mentioned methods enhance the encoder by pre-training or the multi-task learning paradigm. Instead, we propose an independent cross-encoder as the schema item classifier which is easier to be trained. We use the classifier to re-organize the input of the seq2seq model, which can produce a more direct impact on schema linking. \citet{ben2019global} calculate a relevance score for each schema item, which is then used as the soft coefficient of the schema items in the subsequent graph encoder. Compared with them, our method can be viewed as a hard filtering of schema items which can reduce noise more effectively.

\subsection{Intermediate Representation}
Because there is a huge gap between natural language questions and their corresponding SQL queries, some works have focused on how to design an efficient intermediate representation (IR) to bridge the aforementioned gap~\citep{tao2018syntaxsqlnet, jiaqi2019towards, yujian2021naturalsql}. Instead of directly generating full-fledged SQL queries, these IR-based methods encourage models to generate IRs, which can be translated to SQL queries via a non-trainable transpiler.

\section{Conclusion}
In this paper, we propose \model, a simple yet powerful Text-to-SQL parser. 
We first train a cross-encoder to rank and filter schema items which are then injected into the encoder of the seq2seq model. We also let the decoder generate the SQL skeleton first, which can implicitly guide the subsequent SQL generation. To a certain extent, such a framework decouples schema linking and skeleton parsing, which can alleviate the difficulty of Text-to-SQL.
Extensive experiments on Spider and its three variants demonstrate the performance and robustness of \model.

\section*{Acknowledgments}
We thank Hongjin Su and Tao Yu for their efforts in evaluating our model on Spider's test set. We also thank the anonymous reviewers for their helpful suggestions. This work is supported by National Natural Science Foundation of China (62076245, 62072460, 62172424, 62276270) and Beijing Natural Science Foundation (4212022).

\bibliography{aaai23}

\end{document}